\documentclass[sigconf, nonacm]{acmart}

\usepackage{booktabs}
\usepackage{multirow}
\usepackage{graphicx}
\usepackage{xcolor}
\usepackage{amsmath}
\usepackage{microtype}

\setcopyright{none}
\renewcommand\footnotetextcopyrightpermission[1]{}
\acmConference[RLEval @ ACM CAIS 2026]{RLEval: Methods and RL Environments for Evaluating AI Agents}{May 26, 2026}{San Jose, CA}

\title{Counterfactual Evaluation Reveals Hidden Capability Profiles \\ in Clinical LLMs and Agents}

\author{Matt Turk}
\affiliation{%
  \institution{Protege Data Lab}
  \city{New York City}
  \country{USA}
}
\email{matt.turk@withprotege.ai}

\begin{abstract}
Two clinical AI systems can score nearly identically on coverage-based rubrics yet behave radically differently when their patient inputs change: one updates its recommendations to match the new clinical signal, the other produces the same output regardless. Standard evaluation cannot tell them apart. We introduce the \textbf{Causal Sensitivity Score (CSS)}, a pre-registered interventional metric that mutates oncology tumor-board cases along five clinically meaningful dimensions (biomarker flips, prior-treatment failures, biomarker strips, surgery-status changes, stage perturbations) and scores in $\{0, 0.5, 1.0\}$ whether each model's recommendations update in the pre-registered correct direction. Benchmarked against the published Consensus Match Score (CMS), a coverage-based weighted recall, six frontier models from three labs in single-shot inference on 224 cases rank in nearly opposite orders on the two metrics: all six change rank, the CMS-worst model becomes CSS-best, and one model that is upper-mid on CMS is dead last on CSS. We further surface a universal safety blind spot under our pre-registered scoring rule: every frontier model fails on surgery-status interventions ($\le 17.2\%$ CSS on Family D), a finding CMS does not expose. The metric transfers directly to tool-using agents: a ReAct-style experiment shows tool use lifts CSS for five of six models ($+2.5$ to $+20.3$pp), yet the lowest-CSS model retrieves the same chart sections as the others and still does not update its recommendations, suggestive of a structural-responsiveness deficit visible only under counterfactual evaluation. Cross-judge replication and three-rater medical-professional validation confirm the aggregate findings. Interventional pre-registered metrics like CSS complement coverage-based evaluation for clinical AI agents: they capture responsiveness signal coverage cannot, and offer a candidate dense reward for future agentic RL.
\end{abstract}

\keywords{Counterfactual evaluation, causal sensitivity, agent evaluation, RL reward signals, LLM-as-judge, clinical AI}

\begin{document}
\maketitle

\section{Introduction}
\label{sec:intro}

LLMs and LLM-powered agents are increasingly deployed in clinical AI (treatment recommendation, triage, tumor-board summarization), where evaluation determines whether they ship. The dominant paradigm scores outputs against reference behaviors via string similarity or LLM-as-judge rubrics~\cite{zheng2024judge,liang2023helm}. Both ask: \emph{does the output look right?} Neither asks: \emph{is the model updating its output for the right reasons?} An oncology AI that proposes FOLFIRINOX for a pancreatic case scores equally well on coverage-based metrics whether the patient is treatment-naive or whether it just always proposes FOLFIRINOX for pancreatic cases.

\textbf{Why this matters for agent evaluation.} Frontier deployments increasingly run as tool-using agents that fetch patient information themselves. Coverage-based metrics on agent outputs face a sharper look-right-vs.-be-right problem: an agent can make many tool calls, retrieve the right information, and still produce a recommendation that ignores what it found. Interventional metrics are the natural fit because they grade \emph{behavioral responsiveness}: did the agent's output update appropriately when its tool returns were changed?

We introduce the \textbf{Causal Sensitivity Score (CSS)}, a pre-registered interventional metric. For each intervention (flipping HER2 status, injecting a prior failed therapy, removing biomarker mentions, toggling surgery status, etc.), CSS scores in $\{0, 0.5, 1.0\}$ whether recommendations update in the pre-registered correct direction. We evaluate six frontier models from three labs (OpenAI \texttt{gpt-5}, \texttt{gpt-5.4}, \texttt{gpt-5.4-mini}; Anthropic \texttt{claude-opus-4-7}, \texttt{claude-sonnet-4-6}; xAI \texttt{grok-4.20-0309-reasoning}) in two settings: single-shot LLM inference on all 224 expert-annotated tumor-board cases (\S\ref{sec:results}), and a tool-using ReAct agent~\cite{yao2023react} on a 100-tuple Family D subset where interventions propagate through tool returns rather than the prompt (\S\ref{sec:agentic}).

\textbf{Our findings:}
\begin{itemize}
  \item \textbf{Rank reversal (single-shot).} CMS and CSS rank the six models in nearly opposite orders ($\rho = -0.49$; all six change rank); the CMS-worst model is CSS-best, and \texttt{gpt-5.4} is upper-mid on CMS (rank 4 of 6) but dead last on CSS. All six models fail Family D (surgery status) at $\le 17.2\%$ under the pre-registered scoring rule, a universal failure CMS does not expose.
  \item \textbf{Agent transfer.} CSS transfers to tool-using agents without modification. Tool use lifts CSS for five of six models on Family D ($+2.5$ to $+20.3$pp); \texttt{gpt-5.4} alone is essentially unchanged despite retrieving the same chart sections as the responsive five, suggestive of structural responsiveness rather than information access.
  \item \textbf{Validation.} Cross-judge replication (uniform Opus) preserves rank order ($\rho=+1.00$); three-rater medical-professional annotation on a 100-tuple subset confirms aggregate per-family failure rates (Family D: LLM mean $0.10$ vs.\ human mean $0.09$).
\end{itemize}

\section{Method}
\label{sec:method}

\subsection{Pre-registered Intervention Catalog}

We curate 12 interventions across five clinically motivated families (Table~\ref{tab:catalog}). Each is specified in a YAML catalog with five fields committed \emph{before} any model is evaluated: applicability filter, mutation rule (regex \texttt{replace}/\texttt{delete}/\texttt{insert}), pre-registered expected output change, $\{0,0.5,1.0\}$ scoring rule, and family label A--E (full schema in App~\ref{app:catalog}).

\begin{table}[t]
\caption{Pre-registered intervention families. ``Eligible'' counts the tuples that pass the catalog applicability filter; the per-family scored $n$ in Table~\ref{tab:perfamily} excludes regex no-op mutations (App~\ref{app:impl}), e.g., 73/153 eligible Family C tuples produced no-ops and are dropped.}
\label{tab:catalog}
\small
\begin{tabular}{@{}lll@{}}
\toprule
Family & Mutation type & Eligible $n$ \\
\midrule
A: biomarker flip       & replace HER2/ER/PD-L1 status & 129 \\
B: prior treatment      & inject prior-line progression & 269 \\
C: biomarker strip      & delete biomarker mentions & 153 \\
D: surgery status       & toggle resection history & 306 \\
E: stage perturbation   & change disease stage & 5 \\
\bottomrule
\end{tabular}
\end{table}

Pre-registration rules out post-hoc family selection and the ``metric designed to fit the result'' critique. The catalog and scoring rules were authored by the author and have not yet undergone independent clinical vetting (App~\ref{app:cameraready}).

\subsection{Causal Sensitivity Score}

For each (model $m$, intervention $i$, case $c$) where $i$ applies, we generate baseline recommendations from the unmodified packet and intervened recommendations from the mutated packet. A judge LLM receives both, the pre-registered expected change, and the scoring rule, and emits $\{0.0, 0.5, 1.0\}$ for (no change / acknowledged but unchanged / updated correctly). Each tuple is processed through a two-stage pipeline (case summary $\rightarrow$ recommendations); the judge sees only the recommendations. CSS is the mean score across $(c, i)$ tuples, in aggregate and per-family.

\textbf{Self-judging avoidance.} We use \texttt{gpt-5.4} as the default judge with \texttt{claude-opus-4-7} as the judge when \texttt{gpt-5.4} is the model under test, consistent with prior self-preference findings~\cite{zheng2024judge}. \S\ref{sec:judge_sensitivity} reports a uniform-Opus replication.

\subsection{Generalization to Tool-Using Agents}
\label{sec:generalization}

CSS requires only (a) inputs that admit pre-registered counterfactual mutations and (b) a pre-specifiable correct output-update direction. Both transfer directly to tool-using agents, where mutations can be applied to tool returns (e.g., flip a knowledge-base retrieval), to planning state, or to environment observations. The single-shot setting we report on is the cleanest experimental control; \S\ref{sec:agentic} runs the same protocol against ReAct agents and shows the metric and findings transfer.

\section{Experimental Setup}
\label{sec:setup}

\textbf{Cohort.} 224 oncology tumor-board cases, each with a chronological patient packet (median $\sim$80k characters) and ground-truth treatment recommendations from a two-round expert oncologist consensus protocol. Treatments are labeled \texttt{strong} (clear consensus), \texttt{tacit} (tacit consensus), \texttt{mixed} (mixed evidence), or \texttt{refusal} (clear rejection).

\textbf{Models.} The six frontier models listed in \S\ref{sec:intro}, called via official APIs at default temperature.

\textbf{Comparison metric: Consensus Match Score (CMS).} A published weighted recall against the oncologist-consensus treatment list:
\[
\text{CMS} = 0.6 R_{\text{strong}} + 0.2 R_{\text{tacit}} + 0.15 (1 - V_{\text{refusal}}) + 0.05 P_{\text{extra}},
\]
with $R$ recalls of strong/tacit-consensus treatments, $V_{\text{refusal}}$ the rate of recommending a rejected treatment, and $P_{\text{extra}}$ judge-rated plausibility of off-list recommendations. CMS measures \emph{output coverage} (does the recommendation overlap the consensus?); CSS, by contrast, measures \emph{input responsiveness}.

\section{Results}
\label{sec:results}

\subsection{Rank Disagreement Between CMS and CSS}

\begin{figure}[t]
\centering
\includegraphics[width=\linewidth]{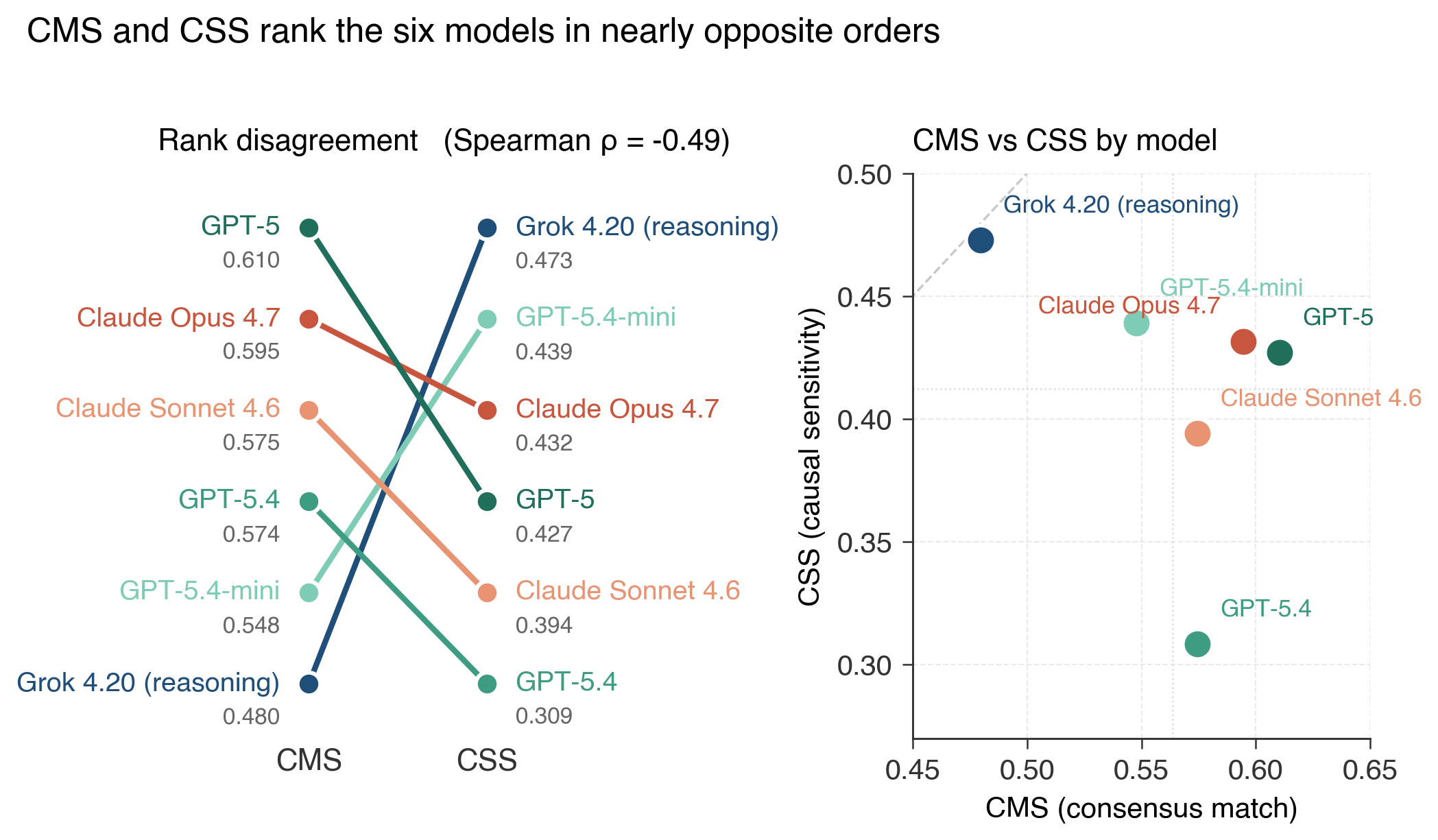}
\caption{Rank disagreement between CMS and CSS for six frontier models from three labs. Spearman $\rho = -0.49$ (exact permutation $p=0.36$ at $n=6$, underpowered); all six models change rank between the two metrics. The CMS-worst model (\texttt{grok-4.20-reasoning}) is CSS-best; the CMS-best (\texttt{gpt-5}) is fourth on CSS.}
\label{fig:headline}
\end{figure}

\begin{table}[t]
\caption{Headline: CMS vs.\ CSS for six frontier models. The two metrics rank the models in nearly opposite orders.}
\label{tab:headline}
\small
\begin{tabular}{@{}lcccc@{}}
\toprule
Model & CMS & rank & CSS & rank \\
\midrule
\texttt{gpt-5}                    & \textbf{0.610} & 1 & 0.427 & 4 \\
\texttt{claude-opus-4-7}          & 0.595 & 2 & 0.432 & 3 \\
\texttt{claude-sonnet-4-6}        & 0.575 & 3 & 0.394 & 5 \\
\texttt{gpt-5.4}                  & 0.575 & 4 & \textbf{0.309} & \textbf{6} \\
\texttt{gpt-5.4-mini}             & 0.548 & 5 & 0.439 & 2 \\
\texttt{grok-4.20-reasoning}      & \textbf{0.480} & \textbf{6} & \textbf{0.473} & \textbf{1} \\
\bottomrule
\end{tabular}
\end{table}

Table~\ref{tab:headline} and Figure~\ref{fig:headline} show the headline. The six models cluster within $13.1$pp on CMS ($0.480$--$0.610$) but span $16.4$pp on CSS ($0.309$--$0.473$); Spearman $\rho = -0.49$ (exact permutation $p=0.36$ at $n=6$, underpowered). \emph{All six} models change rank: the most striking flip is \texttt{grok-4.20-reasoning} (CMS 6 $\rightarrow$ CSS 1) and \texttt{gpt-5} (CMS 1 $\rightarrow$ CSS 4). We treat the rank disagreement as a descriptive pattern; adding models will sharpen the inferential claim.

\subsection{Per-Family Capability Profiles}

\begin{figure}[t]
\centering
\includegraphics[width=\linewidth]{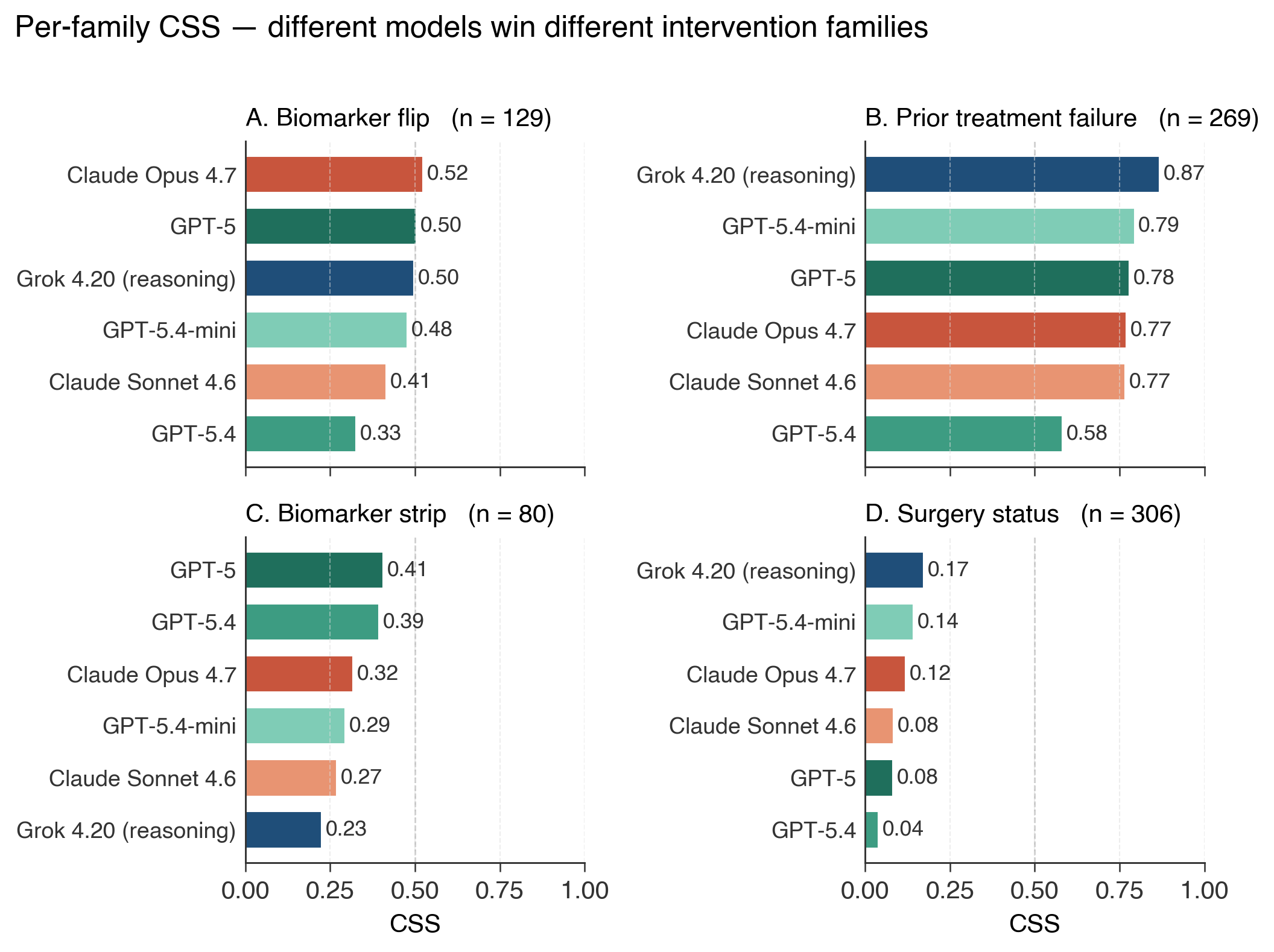}
\caption{Per-family CSS for six frontier models, small-multiples view. Different models win different families: \texttt{claude-opus-4-7} on biomarker recognition (A), \texttt{gpt-5} on biomarker stripping (C), \texttt{grok-4.20} on prior-treatment (B) and surgery status (D). All six models fail catastrophically on Family D.}
\label{fig:perfamily}
\end{figure}

\begin{table*}[t]
\caption{Per-family CSS, six models, five intervention families. Bold = winner per row.}
\label{tab:perfamily}
\small
\begin{tabular}{@{}lcccccc@{}}
\toprule
Family & opus-4-7 & sonnet-4-6 & gpt-5 & gpt-5.4 & gpt-5.4-mini & grok-4.20 \\
\midrule
A: biomarker flip          ($n{=}129$) & \textbf{0.523} & 0.415 & 0.504 & 0.326 & 0.477 & 0.496 \\
B: prior treatment failure ($n{=}269$) & 0.770 & 0.766 & 0.779 & 0.582 & 0.794 & \textbf{0.868} \\
C: biomarker strip         ($n{=}80$)  & 0.316 & 0.269 & \textbf{0.406} & 0.394 & 0.294 & 0.225 \\
D: surgery status          ($n{=}306$) & 0.119 & 0.083 & 0.082 & 0.039 & 0.142 & \textbf{0.172} \\
\midrule
\textbf{Aggregate} & 0.432 & 0.394 & 0.427 & 0.309 & 0.439 & \textbf{0.473} \\
\bottomrule
\end{tabular}
\end{table*}

Table~\ref{tab:perfamily} decomposes CSS by intervention family. Different models win different families: \texttt{claude-opus-4-7} on biomarker flips (A); \texttt{gpt-5} on biomarker stripping (C); \texttt{grok-4.20} on prior-treatment failure (B) and surgery status (D). \texttt{gpt-5.4} is dead last on A, B, and D; on C it is second (behind \texttt{gpt-5}). This per-family decomposition reveals capability profiles aggregate metrics destroy. (Family E, stage-perturbation, has only $n=5$ eligible cases; we do not draw conclusions from it, App~\ref{app:familyE}.)

\subsection{Universal Failure Mode: Family D}

The strongest model on surgery-status interventions (\texttt{grok-4.20}) scores 17.2\%; the weakest (\texttt{gpt-5.4}) scores 3.9\%. \emph{Every} frontier model from \emph{every} lab fails to update treatment recommendations correctly when surgery status flips. This is a clinically meaningful safety finding (treatment timing depends entirely on whether the patient was resected) that CMS does not surface, because CMS only checks recommendation overlap with consensus, not behavioral change under a counterfactual.

\subsection{Score-Distribution Diagnostics}

\begin{table}[t]
\caption{Per-model score distribution. ``Wrong'' = $0.0$, ``Partial'' = $0.5$, ``Correct'' = $1.0$.}
\label{tab:distribution}
\small
\begin{tabular}{@{}lccc@{}}
\toprule
Model & Wrong & Partial & Correct \\
\midrule
\texttt{grok-4.20-reasoning}   & \textbf{40.8\%} & 23.8\% & 35.4\% \\
\texttt{gpt-5}                 & 50.6\% & 13.4\% & \textbf{36.0\%} \\
\texttt{gpt-5.4-mini}          & 45.2\% & 21.7\% & 33.1\% \\
\texttt{claude-opus-4-7}       & 47.8\% & 18.0\% & 34.1\% \\
\texttt{claude-sonnet-4-6}     & 54.0\% & 13.2\% & 32.8\% \\
\texttt{gpt-5.4}               & \textbf{59.6\%} & 19.1\% & \textbf{21.3\%} \\
\bottomrule
\end{tabular}
\end{table}

The score distribution (Table~\ref{tab:distribution}) concretizes the two failure modes: \texttt{gpt-5.4} sits at 60\% wrong-direction / 21\% correct (the ``looks fine on CMS, structurally less responsive'' case); \texttt{grok-4.20-reasoning} is the inverse, with the lowest wrong-direction rate (40.8\%) and second-highest correct rate (35.4\%).

\subsection{Judge Sensitivity (Cross-Judge Replication)}
\label{sec:judge_sensitivity}

To rule out the asymmetric judge dispatch as a confound, we re-judge every tuple with \texttt{claude-opus-4-7} as a single judge for \emph{all} models on the same 4{,}727 tuples. Rank order is identical under both configurations (Spearman $\rho=+1.00$); per-model inter-judge $\kappa$ on $\{0,0.5,1.0\}$ is $0.61$--$0.69$ for the five cross-judged models. Opus is a \emph{stricter} judge than gpt-5.4: aggregate CSS drops $4$--$7$pp under Opus for the five non-gpt-5.4 models. Since \texttt{gpt-5.4} was already Opus-judged by default, the original asymmetric dispatch was biased \emph{against} \texttt{gpt-5.4} (held to a stricter standard than the other five, which were judged by the more lenient gpt-5.4), so its persistent last-place ranking is the harder of the two directions to obtain by chance. \texttt{gpt-5.4} still ranks last under uniform Opus judging; the deficit is not a judge-dispatch artifact (full table in App~\ref{app:judge}). Construct validity holds across hard-case, latent, and cancer-category strata (App~\ref{app:stratification}).

\subsection{Generalization Experiment: Tool-Using Agent}
\label{sec:agentic}

\begin{figure}[t]
\centering
\includegraphics[width=\linewidth]{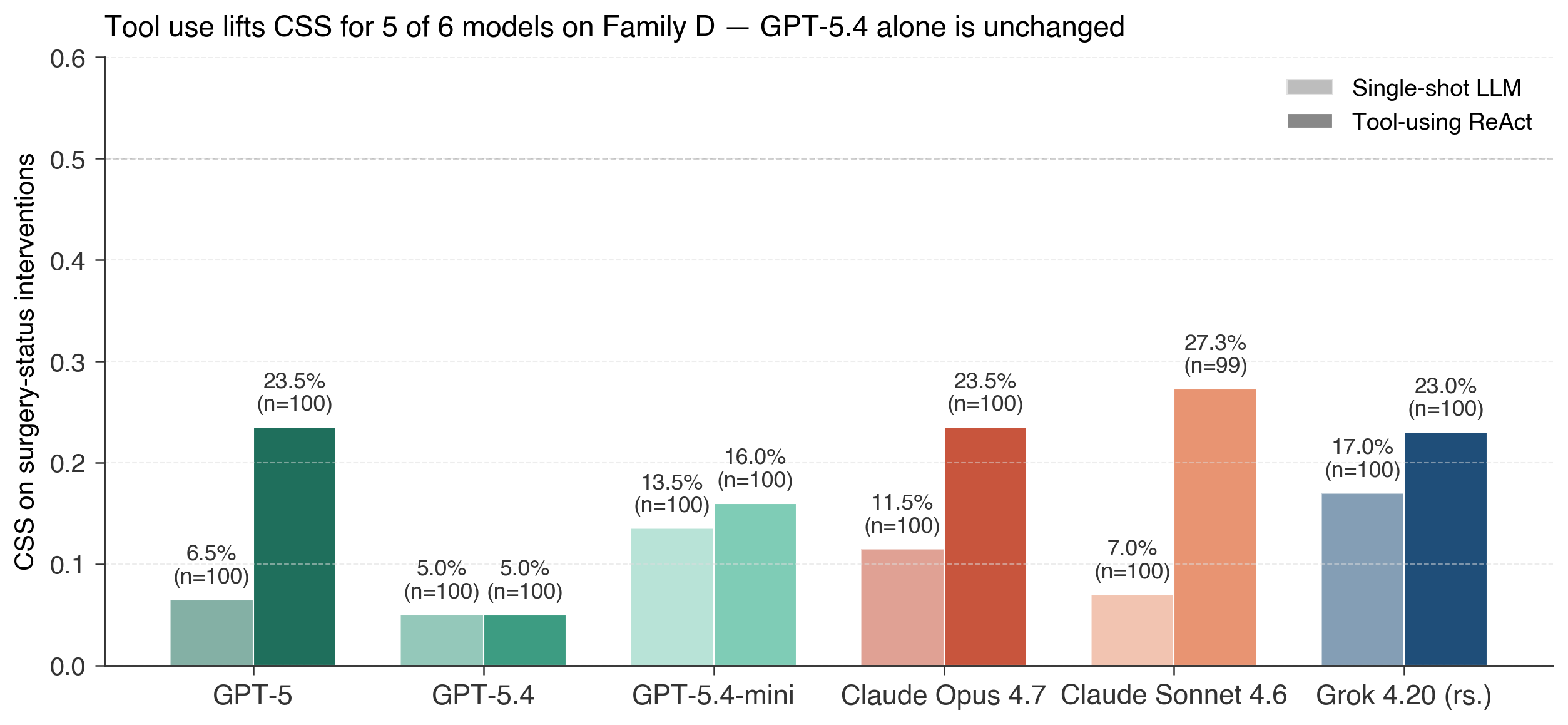}
\caption{Tool use lifts CSS on Family D for 5 of 6 frontier models (gain $+2.5$ to $+20.3$pp). \texttt{gpt-5.4} is essentially unchanged ($0.050\rightarrow 0.050$), consistent with a structural responsiveness deficit rather than information-access failure. Up to 100 case-intervention tuples per model (one sonnet row dropped as missing data, $n=99$).}
\label{fig:agentic}
\end{figure}

We re-run Family D in a tool-using agent setting: the agent has no chart in context and a single \texttt{read\_chart\_section(section)} tool over ten chart sections (demographics, diagnoses, biomarkers, medications, procedures, encounters, labs, vitals, allergies, overview). It investigates ReAct-style~\cite{yao2023react}; interventions mutate the underlying packet so the change propagates through retrieval. Judge and scoring are unchanged. We evaluate all six models on 100 tuples each; all call the tool $7.0$--$8.3$ times per case and query \texttt{procedures} at similar rates; they retrieve the same information.

\emph{Five of six models lift substantially under tool use}, with single-shot$\rightarrow$tool-using CSS gains of $+17.0$pp (\texttt{gpt-5}), $+20.3$pp (\texttt{claude-sonnet-4-6}, rank 4$\rightarrow$1 on Family D), $+12.0$pp (\texttt{claude-opus-4-7}), $+6.0$pp (\texttt{grok-4.20}), $+2.5$pp (\texttt{gpt-5.4-mini}); \texttt{gpt-5.4} is essentially unchanged ($0.050\rightarrow 0.050$). Tool use helps 19--34 of up to 100 tuples for responders, only 9 for \texttt{gpt-5.4} (App~\ref{app:percase}). The asymmetry is substantive: \texttt{gpt-5.4} retrieves the same sections as the responsive five and still does not update, suggestive of structural responsiveness rather than information access. Even the best tool-using model (0.273) sits well below 50\%, so tool use mitigates but does not close the Family D blind spot.

\subsection{Human Validation}
\label{sec:human_validation}

Three medical-professional annotators independently scored 100 stratified (case, intervention, model) tuples on the same $\{0,0.5,1.0\}$ scale, blinded to model identity and LLM judge scores. Pairwise human-human Cohen's $\kappa$ ranges $0.40$--$0.72$; LLM-vs-majority $\kappa = 0.46$ ($69/100$ exact agreement, App~\ref{app:llm_vs_human}). Per-family LLM-vs-majority $\kappa$ is highest on A ($0.67$) and lowest on C ($0.07$); D is $0.16$. Crucially, the \emph{aggregate} CSS rates agree closely: Family D LLM mean $=0.10$ vs.\ human $=0.09$; C is $0.30$ vs.\ $0.33$. Per-row case-level agreement is moderate, so headline claims should be read as population-level rather than per-case reliability.

Annotators flagged 37/100 rows as medically incoherent (29/49 in Family D); the universal Family D blind spot survives restriction to coherent-only rows (App~\ref{app:coherence}).

\section{Limitations}
\label{sec:limitations}

\textbf{Population-level claims.} Per-row LLM-human $\kappa$ for D and C is moderate-to-low ($0.16$, $0.07$); headline claims should be read as population-level CSS properties rather than per-case reliability. \textbf{Regex-based counterfactuals.} Mutations admit three failure modes that score a correctly-refusing model $0.0$ under the pre-registered rule: semantic no-ops (regex changes text but not meaning), incomplete propagation (one chart section changed while others still imply the original fact, which the model may treat as a data-entry error), and medical incoherence (29/49 D-rows). \textbf{Sample size.} $n=6$ models renders the rank-correlation test underpowered ($p=0.36$); the validation subset is $n=100$ ($15$--$49$ per family), so per-family $\kappa$ carries sampling noise. \textbf{Scope.} Findings are specific to oncology tumor-board cases; CSS methodology generalizes elsewhere given a domain-appropriate catalog. \textbf{Agent attribution.} The agent setting introduces variables besides tool use (sectioned retrieval, ReAct prompting, 8 KB cap); the gpt-5.4 persistence is consistent with structural responsiveness, not proof. Catalog provenance and camera-ready / future-work plans are in App~\ref{app:cameraready}.

\section{Related Work}
\label{sec:related}

LLM-as-judge reliability is established in Zheng et al.~\cite{zheng2024judge}; counterfactual probing of LLM reasoning in Saparov \& He~\cite{saparov2023greedy}; holistic and agent-eval aggregation in HELM~\cite{liang2023helm} and AgentBench~\cite{liu2023agentbench}; clinical LLM QA evaluation in Singhal et al.~\cite{singhal2023expertmed}; step-level process rewards in Lightman et al.~\cite{lightman2023letsverify}. CSS extends interventional probing to agentic clinical outputs, decomposes by intervention family, and operates as a candidate dense reward in the spirit of process supervision.

\section{Conclusion}

We introduce the Causal Sensitivity Score, a pre-registered counterfactual metric, and apply it to six frontier models on 224 oncology cases (single-shot) and a 100-tuple Family D subset (tool-using ReAct), benchmarked against the Consensus Match Score (CMS) coverage metric. Under the pre-registered scoring rule, CMS and CSS rank the models in nearly opposite orders: all six change rank, the CMS-worst model becomes CSS-best, and \texttt{gpt-5.4} sits at CMS rank 4 of 6 but is dead last on CSS. Every model fails Family D surgery-status interventions at $\le 17.2\%$, a universal safety blind spot CMS does not expose. In the agent setting, tool use lifts CSS for five of six models ($+2.5$ to $+20.3$pp); \texttt{gpt-5.4} alone is essentially unchanged despite retrieving the same chart sections as the responsive five, a pattern consistent with structural responsiveness rather than information access. Cross-judge replication and three-rater medical-professional validation support the aggregate findings.

We see CSS as a complement to CMS rather than a replacement: it captures input-responsiveness signal coverage-based metrics cannot, transfers without modification from single-shot LLMs to tool-using agents, and offers a candidate dense reward for future agentic RL experiments. Together, these results indicate that the rank disagreement between CMS and CSS, the universal Family D failure, and the gpt-5.4 retrieval-without-update pattern are robust to judge choice and to human adjudication on the validation subset. Whether a clinical AI agent's recommendations update appropriately when the patient's facts change is an evaluation question worth measuring directly, and one that coverage-based metrics cannot answer.

\bibliographystyle{plain}
\bibliography{references}

\begin{acks}
We thank Engy Ziedan and Wes Hopkins of the Protege Data Lab team for their review and assistance in producing and validating these results, and the medical-professional annotators who graded the human-validation subset.
\end{acks}

\appendix

\section{Pre-Registered Intervention Catalog (Excerpt)}
\label{app:catalog}

Each entry in the catalog has five fields, all committed before any model is evaluated. We include two representative entries here; the full catalog (12 entries across families A--E) is in the artifact release.

\begin{verbatim}
- id: A2_HER2_positive_to_negative
  family: A_biomarker_flip
  applicability:
    cancer_type: ["breast", "gastric"]
    requires_match_in_packet:
      "HER2[\\s-]?(positive|\\+)"
  mutation_type: replace
  mutation_pattern:
      "HER2[\\s-]?(positive|\\+)"
  mutation_replacement: "HER2-negative"
  expected_change:
    must_drop: ["trastuzumab", "T-DXd",
                "pertuzumab", "tucatinib"]
  scoring_rule:
    1.0: "all HER2-targeted recs dropped"
    0.5: "some dropped, or hedged in
          rationale only"
    0.0: "HER2-targeted recs unchanged"

- id: D1_remove_surgery_history
  family: D_surgery_status
  applicability:
    requires_match_in_packet:
      "\\[Procedure\\][^\\n]*?(resection|
       craniotomy|lumpectomy|mastectomy|
       gastrectomy|whipple|hepatectomy)"
  mutation_type: delete
  mutation_pattern:
      "\\[Procedure\\][^\\n]*?(resection|
       ...)[\\s\\S]*?(?=\\n---|\\n\\[)"
  expected_change:
    timing_must_shift:
      - adjuvant -> primary
      - post-operative -> definitive
    rationale_must_mention: "resection
        no longer documented"
  scoring_rule:
    1.0: "timing shifted as specified"
    0.5: "rationale acknowledges but
          recommendations unchanged"
    0.0: "no change"
\end{verbatim}

\section{Sub-Population Stratification}
\label{app:stratification}

CSS by latent-consensus flag and by cancer-category folder, six models. The gpt-5.4 deficit persists across all strata. Grok leads in 4 of 5 strata; in the mixed-evidence stratum, \texttt{gpt-5.4-mini} narrowly edges Grok ($0.529$ vs.\ $0.471$).

\begin{table}[h]
\caption{Sub-population CSS (all six models, all interventions).}
\small
\begin{tabular}{@{}lcccccc@{}}
\toprule
Subset & opus & sonnet & gpt-5 & gpt-5.4 & mini & grok \\
\midrule
Non-latent  & 0.483 & 0.446 & 0.458 & 0.321 & 0.492 & 0.500 \\
Latent      & 0.422 & 0.385 & 0.422 & 0.306 & 0.430 & 0.468 \\
\midrule
Clear cons. & 0.434 & 0.383 & 0.425 & 0.317 & 0.434 & 0.471 \\
Tacit cons. & 0.426 & 0.413 & 0.429 & 0.293 & 0.442 & 0.478 \\
Mixed evid. & 0.441 & 0.441 & 0.441 & 0.324 & 0.529 & 0.471 \\
\bottomrule
\end{tabular}
\end{table}

\section{Tool-Using Agent: Per-Case Effects}
\label{app:percase}

Per-model breakdown of how tool use changes the score on each of the 100 matched (case, intervention) tuples on Family D.

\begin{table}[h]
\caption{Per-case effect of tool use on Family D. ``Help'' = tool-using CSS strictly higher; ``Hurt'' = tool-using CSS strictly lower; ``Same'' = unchanged.}
\small
\begin{tabular}{@{}lccc@{}}
\toprule
Model & Help & Same & Hurt \\
\midrule
\texttt{claude-sonnet-4-6}      & 34 & 64 & 1 \\
\texttt{gpt-5}                  & 33 & 64 & 3 \\
\texttt{claude-opus-4-7}        & 28 & 66 & 6 \\
\texttt{grok-4.20-reasoning}    & 20 & 73 & 7 \\
\texttt{gpt-5.4-mini}           & 19 & 68 & 13 \\
\texttt{gpt-5.4}                & 9  & 85 & 6 \\
\bottomrule
\end{tabular}
\end{table}

\begin{figure}[h]
\centering
\includegraphics[width=\linewidth]{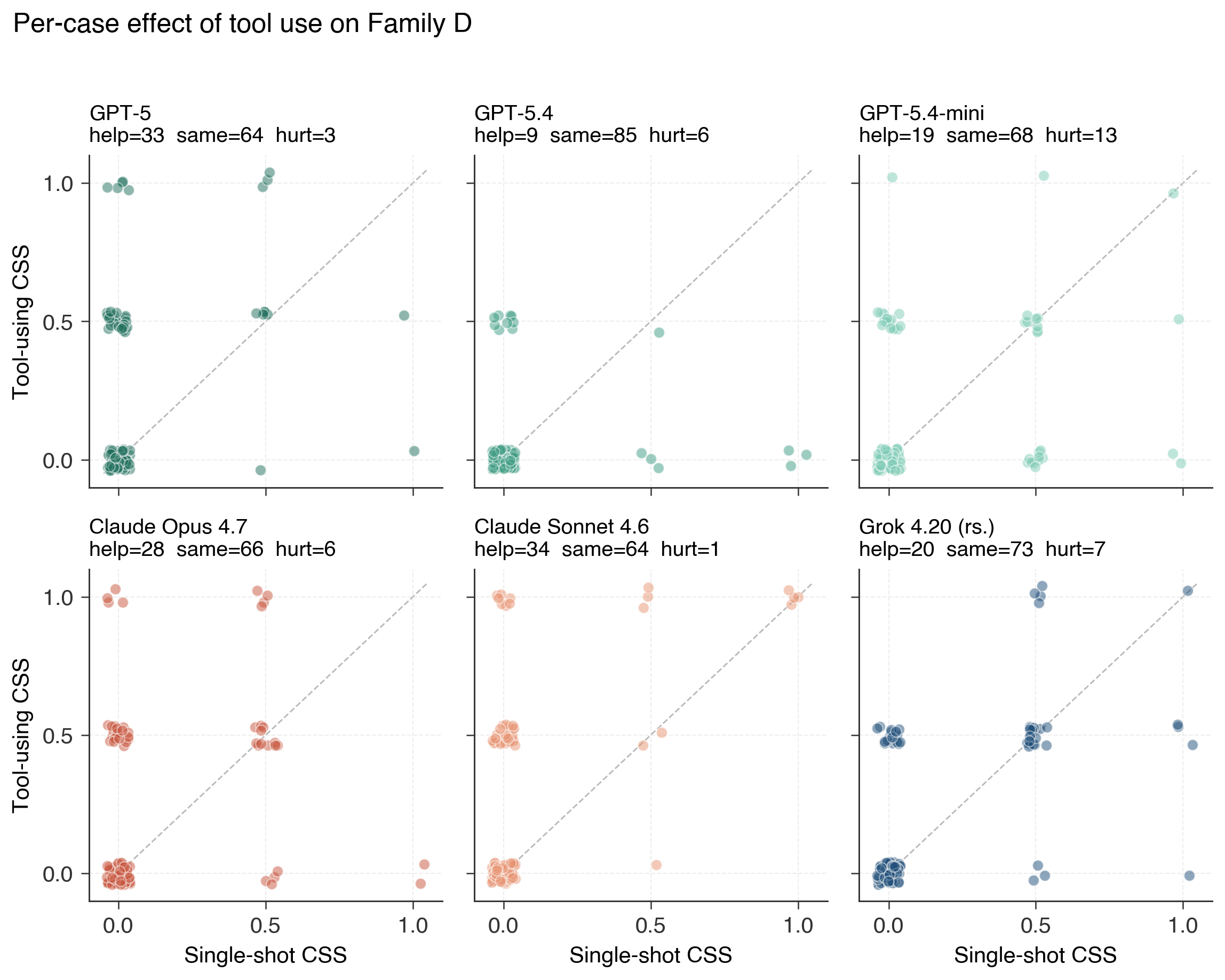}
\caption{Per-case effect of tool use. Each point is one (case, intervention) tuple; x-axis is single-shot CSS, y-axis is tool-using CSS. Points above the diagonal are tuples where tool use helped. The mass of points above the diagonal for five models, and the near-diagonal cluster for \texttt{gpt-5.4}, makes the asymmetry visible at the per-case level.}
\label{fig:percase}
\end{figure}

\section{Tool-Using Agent: Score Distribution Shift}
\label{app:agentic_dist}

Wrong / partial / correct breakdown for the agentic experiment, single-shot LLM (top half) vs.\ tool-using ReAct (bottom half), six models. The wrong-direction rate drops $6$--$30$pp for the five responsive models; only $3$pp for \texttt{gpt-5.4}.

\begin{table}[h]
\small
\begin{tabular}{@{}lccc@{}}
\toprule
Model & Wrong & Partial & Correct \\
\midrule
\multicolumn{4}{l}{\emph{Single-shot LLM (matched 100 tuples)}} \\
\texttt{gpt-5}                  & 89\% & 9\%  & 2\%  \\
\texttt{gpt-5.4}                & 93\% & 4\%  & 3\%  \\
\texttt{gpt-5.4-mini}           & 77\% & 19\% & 4\%  \\
\texttt{claude-opus-4-7}        & 79\% & 19\% & 2\%  \\
\texttt{claude-sonnet-4-6}      & 90\% & 6\%  & 4\%  \\
\texttt{grok-4.20-reasoning}    & 71\% & 24\% & 5\%  \\
\midrule
\multicolumn{4}{l}{\emph{Tool-using ReAct (matched 100 tuples)}} \\
\texttt{gpt-5}                  & 61\% & 31\% & 8\%  \\
\texttt{gpt-5.4}                & 90\% & 10\% & 0\%  \\
\texttt{gpt-5.4-mini}           & 71\% & 26\% & 3\%  \\
\texttt{claude-opus-4-7}        & 61\% & 31\% & 8\%  \\
\texttt{claude-sonnet-4-6}      & 60\% & 26\% & 14\% \\
\texttt{grok-4.20-reasoning}    & 59\% & 36\% & 5\%  \\
\bottomrule
\end{tabular}
\end{table}

\begin{figure}[h]
\centering
\includegraphics[width=\linewidth]{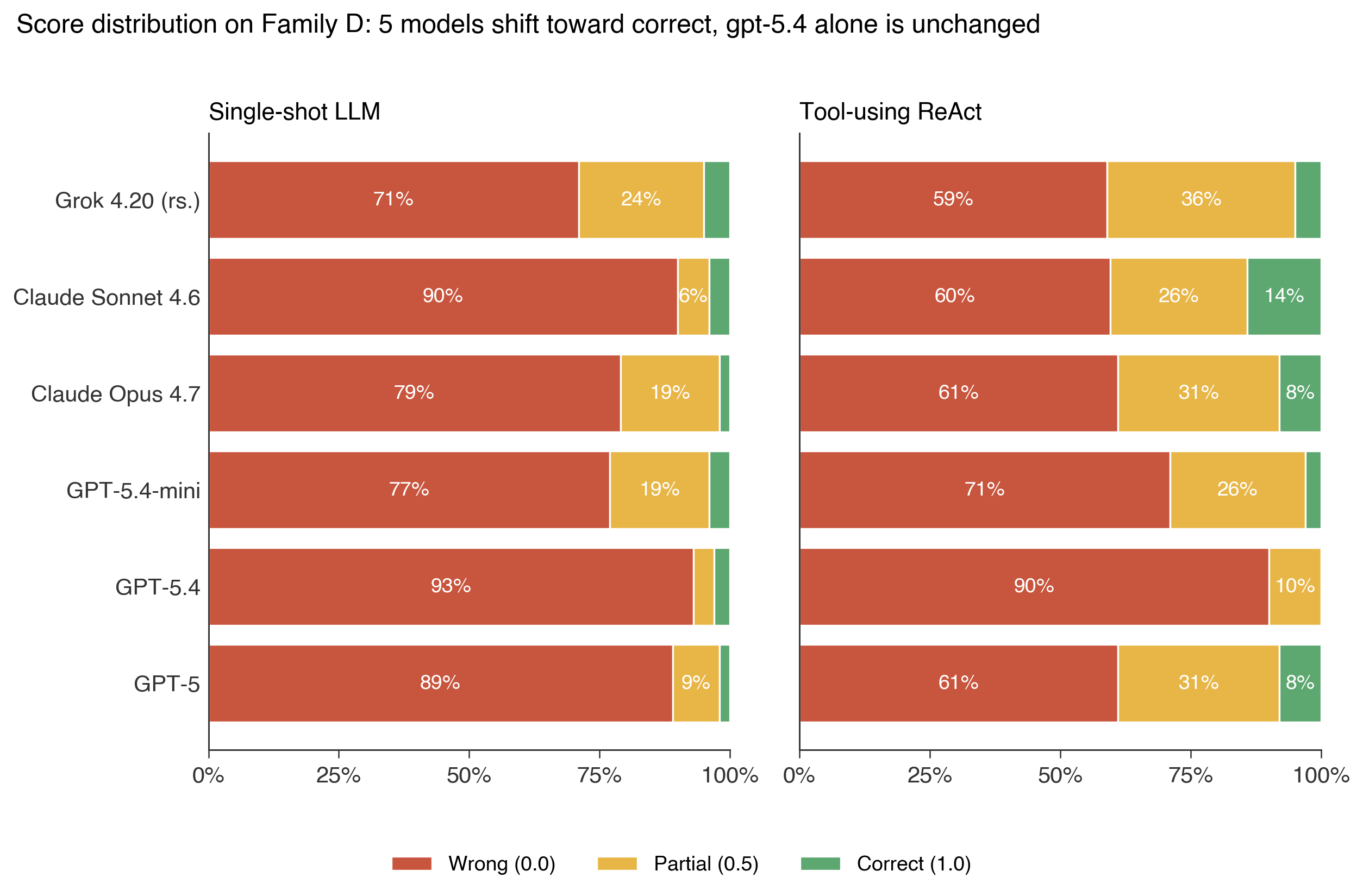}
\caption{Score-distribution shift under tool use. Stacked bars show wrong/partial/correct breakdown for each model in single-shot (left) vs.\ tool-using (right) settings on Family D. The wrong-direction rate drops $6$--$30$pp for the five responsive models; \texttt{gpt-5.4} barely moves.}
\label{fig:agentic_dist}
\end{figure}

\section{Tool-Use Pattern}
\label{app:toolpattern}

\begin{figure}[h]
\centering
\includegraphics[width=\linewidth]{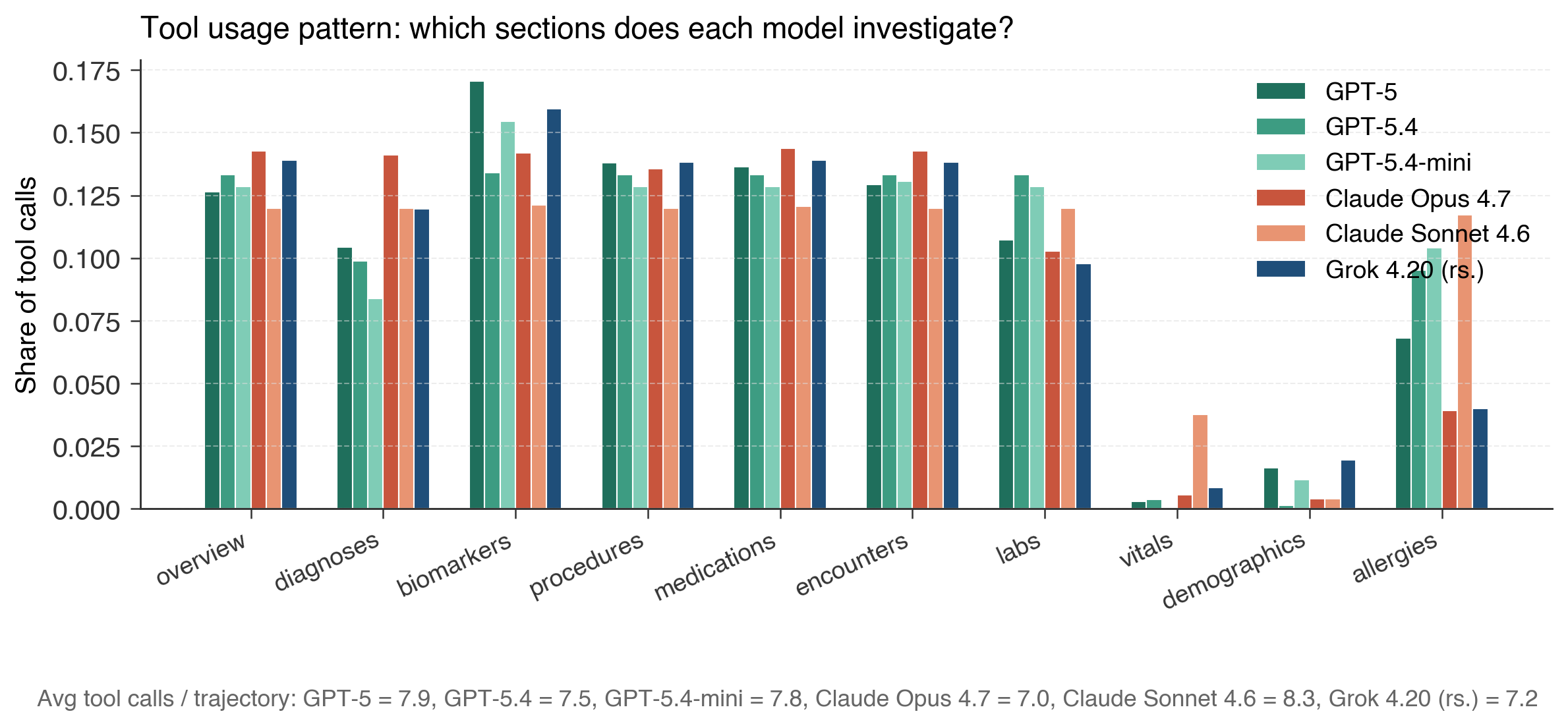}
\caption{Tool-call distribution across the ten chart sections, averaged across 100 case-intervention tuples per model. All six models call the tool $7.0$--$8.3$ times per case and query \texttt{procedures} (where surgery history lives) at similar rates; they retrieve roughly the same information. The CSS asymmetry across models is therefore not explained by retrieval differences.}
\label{fig:toolpattern}
\end{figure}

\section{Compute and Cost}
\label{app:compute}

\begin{itemize}
  \item \textbf{Single-shot baselines.} 6 models $\times$ 224 cases = 1{,}344 baseline inferences (gpt-5 and gpt-5.4-mini reused from prior delivery; remaining 4 models run fresh). Baseline CMS judge: 6 $\times$ 224 = 1{,}344 judge calls.
  \item \textbf{Single-shot interventions.} 6 models $\times$ 789 mutated tuples = 4{,}734 intervention-model inferences. CSS judge: 4{,}734 calls (with self-judge override for \texttt{gpt-5.4} routed to Opus).
  \item \textbf{Tool-using experiment.} 6 models $\times$ (50 baseline cases + 100 intervention tuples) = 900 trajectories, average 7.5 tool calls each, $\approx$ 7{,}600 tool-augmented LLM calls. CSS judge: 600 calls.
\end{itemize}

\section{Implementation Notes}
\label{app:impl}

\textbf{Mutation no-ops.} 73 of 862 eligible (case, intervention) tuples produced no-op mutations (the regex pattern was eligible at the catalog level but did not match the source packet); these are silently dropped from CSS scoring rather than scored as 0.

\textbf{Tool-using agent.} ReAct loop with a 8-call cap (no model hit the cap; mean 7.0--8.3). Chart segmentation is regex-based on tagged entries (\texttt{[Medication]}, \texttt{[Procedure]}, \texttt{[Encounter]}, \texttt{[Lab Results]}, \texttt{[Vitals]}) and on header-delimited blocks (\texttt{PATIENT DEMOGRAPHICS}, \texttt{CONDITIONS}, \texttt{ALLERGIES}). Each section is capped at 8\,KB.

\textbf{Anthropic vs.\ OpenAI tool-call dispatch.} The agent harness has provider-specific paths: OpenAI/xAI use the chat-completions \texttt{tool\_calls} format; Anthropic uses the Messages API \texttt{tool\_use} block format. Both wrap a single shared tool definition, so the agent sees identical tool semantics regardless of provider.

\section{Judge Sensitivity (Cross-Judge Replication) Table}
\label{app:judge}

Per-model CSS under the default judge dispatch (\texttt{gpt-5.4} for five models, Opus override for \texttt{gpt-5.4} as model) versus a uniform Opus-only judge on the same 4{,}727 valid tuples. Rank order is identical under both ($\rho_{\text{judges}}=+1.00$); inter-judge $\kappa$ on $\{0,0.5,1.0\}$ labels is $0.61$--$0.69$ (substantial). Note that \texttt{gpt-5.4}'s row is not a judge-swap: the default dispatch already routed it to Opus, so the $+0.5$pp shift reflects stochastic rerun variability between two Opus passes, not a different judge model.

\begin{table}[h]
\small
\begin{tabular}{@{}lcccc@{}}
\toprule
Model & CSS$_{\text{default}}$ & CSS$_{\text{Opus-only}}$ & $\Delta$ & rank \\
\midrule
\texttt{gpt-5}                     & 0.427 & 0.359 & $-$6.8pp & 4 \\
\texttt{claude-opus-4-7}           & 0.432 & 0.372 & $-$6.0pp & 3 \\
\texttt{gpt-5.4}                   & 0.309 & 0.314 & $+$0.5pp & \textbf{6} \\
\texttt{claude-sonnet-4-6}         & 0.394 & 0.341 & $-$5.3pp & 5 \\
\texttt{gpt-5.4-mini}              & 0.439 & 0.391 & $-$4.8pp & 2 \\
\texttt{grok-4.20-reasoning}       & \textbf{0.473} & \textbf{0.428} & $-$4.4pp & \textbf{1} \\
\bottomrule
\end{tabular}
\end{table}

\section{Agentic Comparison: Single-Shot vs.\ Tool-Using ReAct}
\label{app:agentic_table}

Family D matched 100 case-intervention tuples per model.

\begin{table}[h]
\small
\begin{tabular}{@{}lccc@{}}
\toprule
Model & Single-shot CSS & Tool-using CSS & $\Delta$ \\
\midrule
\texttt{gpt-5}                   & 0.065 & 0.235          & $+$17.0pp \\
\texttt{gpt-5.4}                 & 0.050 & 0.050          & $0.0$pp \\
\texttt{gpt-5.4-mini}            & 0.135 & 0.160          & $+$2.5pp \\
\texttt{claude-opus-4-7}         & 0.115 & 0.235          & $+$12.0pp \\
\texttt{claude-sonnet-4-6}       & 0.070 & \textbf{0.273} & \textbf{$+$20.3pp} \\
\texttt{grok-4.20-reasoning}     & 0.170 & 0.230          & $+$6.0pp \\
\bottomrule
\end{tabular}
\end{table}

\section{Human Validation: Family D Coherence Analysis}
\label{app:coherence}

Of $49$ Family D tuples in the human-annotated subset, $29$ ($59\%$) were flagged by at least one of three annotators as creating a medically incoherent scenario (e.g., curative-intent surgical resection inserted into a metastatic patient's chart). All values below are computed on the $49$-tuple human-annotated subset (\emph{not} the full $306$-tuple Family D); $n_{\text{full}}$ is the number of D-tuples for that model in the subset and $n_{\text{coh}}$ is the subset after removing annotator-flagged rows. Per-model coherent-only CSS is the same or lower than the full-subset CSS for five of six models (\texttt{gpt-5.4-mini} is the exception):

\begin{table}[h]
\small
\begin{tabular}{@{}lccccc@{}}
\toprule
Model & $n_{\text{full}}$ & $n_{\text{coh}}$ & LLM$_{\text{full}}$ & LLM$_{\text{coh}}$ & Human$_{\text{coh}}$ \\
\midrule
\texttt{gpt-5.4-mini}            & 8 & 3 & 0.313 & 0.333 & 0.000 \\
\texttt{gpt-5}                   & 8 & 4 & 0.125 & 0.125 & 0.000 \\
\texttt{grok-4.20-reasoning}     & 8 & 5 & 0.125 & 0.100 & 0.400 \\
\texttt{claude-opus-4-7}         & 9 & 4 & 0.056 & 0.000 & 0.000 \\
\texttt{claude-sonnet-4-6}       & 8 & 1 & 0.000 & 0.000 & 0.000 \\
\texttt{gpt-5.4}                 & 8 & 3 & 0.000 & 0.000 & 0.000 \\
\bottomrule
\end{tabular}
\end{table}

\noindent The catalog issue is concentrated in insert-type mutations (D2 inserts a surgical resection; E1 inserts metastasis) on cases whose disease state is incompatible with the inserted event. For the camera-ready we will tighten D2 and E1 applicability filters with $\lnot$\texttt{metastatic} negative-match guards.

\section{Family E (Stage Perturbation, $n=5$)}
\label{app:familyE}

Reported for completeness; per-family numbers are noisy at $n=5$ and we draw no conclusions. Per-model CSS: \texttt{opus-4-7} 0.800, \texttt{sonnet-4-6} 0.900, \texttt{gpt-5} 1.000, \texttt{gpt-5.4} 0.300, \texttt{gpt-5.4-mini} 0.900, \texttt{grok-4.20-reasoning} 1.000.

\section{LLM Judge vs.\ Human Majority Scatter}
\label{app:llm_vs_human}

\begin{figure}[h]
\centering
\includegraphics[width=\linewidth]{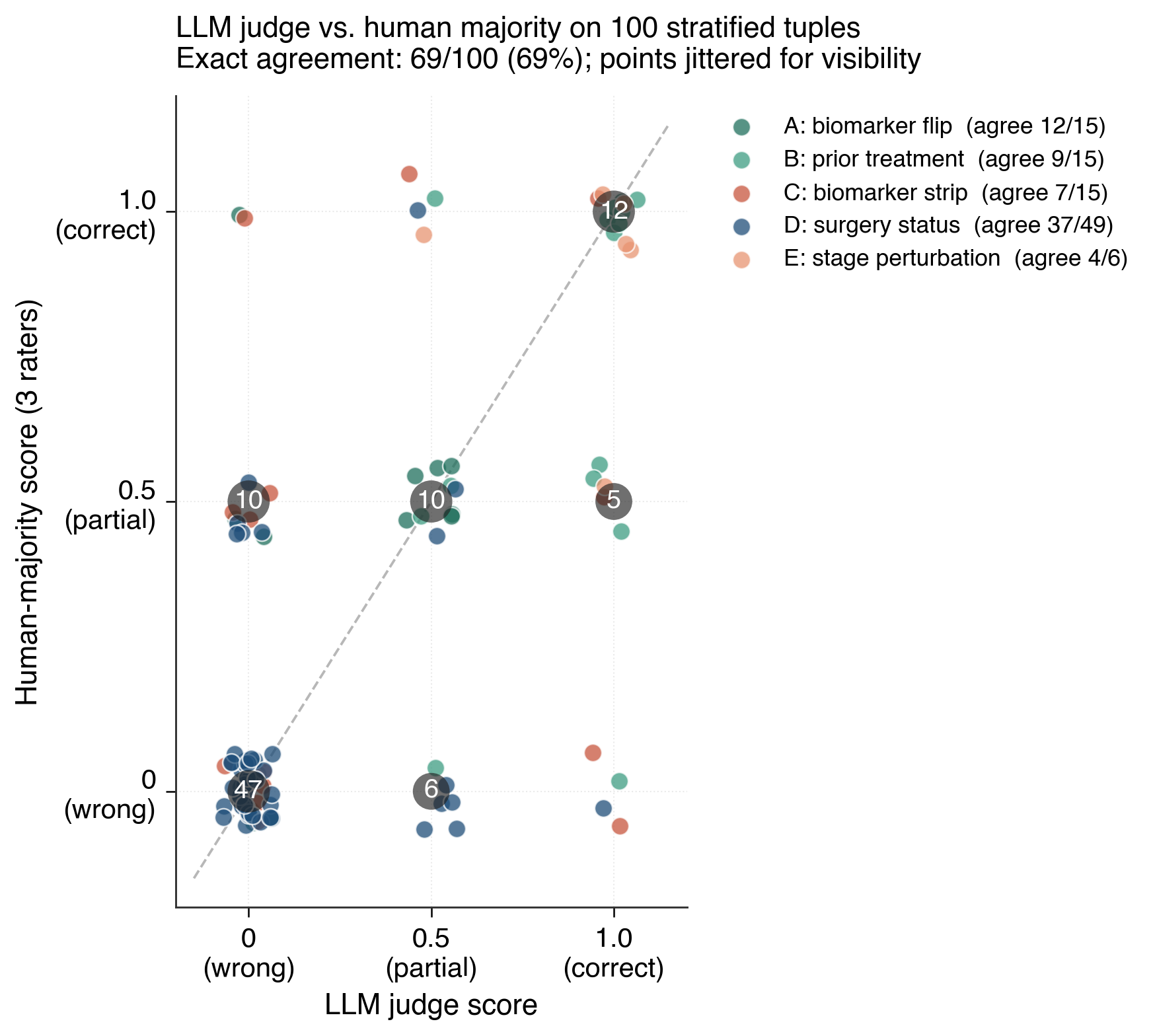}
\caption{Case-level LLM judge vs.\ human-majority score on the 100-tuple validation subset. Points jittered for visibility on the $\{0, 0.5, 1.0\}$ grid; color by intervention family; counts shown in disagreement cells with $\ge 5$ tuples. Exact agreement is $69/100$. Most disagreement clusters at the partial-credit boundary (LLM scores 0.0 where humans score 0.5, and vice versa).}
\end{figure}

\section{Camera-Ready Refinements and Future Work}
\label{app:cameraready}

The workshop camera-ready window is three days. We scope camera-ready refinements to what is achievable in that window and label the rest as future work.

\textbf{Catalog authorship.} The intervention catalog and per-intervention scoring rules were authored by the author and have not yet undergone independent clinical vetting. The rules therefore reflect one researcher's interpretation of what each intervention should change about a model's output, which may differ from a practicing clinician's expectations and may miss clinically meaningful update patterns not enumerated in the catalog. An oncologist review of rule coverage and clinical alignment is a camera-ready item.

\textbf{Camera-ready.} (i) \emph{Tightened insert-filter for D2 and E1.} We will add a $\lnot$\texttt{metastatic} negative-match guard to D2 and E1 applicability filters so curative-intent resection and stage-perturbation insertions are not applied to already-metastatic charts, and re-run the Family D and E subsets on the filtered case set. (ii) \emph{Refusal-credit branch in the scoring rule.} We will add a $0.5$ credit when the model's rationale explicitly identifies the inserted scenario as medically incoherent and leaves recommendations unchanged, and re-judge the affected rows. (iii) \emph{Bootstrap CIs over cases.} We will report case-resampled $95\%$ intervals on per-model CSS and on the rank ordering to characterize stability of the headline reversals.

\textbf{Future work (beyond camera-ready).} (i) \emph{Retrieval-controlled agent baseline} that feeds each model the exact retrieved snippets without the agent loop, to isolate tool-use structural-responsiveness effects from prompt-structure effects. (ii) \emph{Semantic-no-op and incomplete-propagation audits} per mutated chart, requiring a separate validator and a sample-based calibration step. (iii) \emph{Second human-annotation round} with expanded adjudication and refined per-intervention scoring rules to lift per-row $\kappa$ in Families C and D. (iv) \emph{Additional frontier models}, including Gemini and additional reasoning variants. (v) \emph{Bootstrap CIs over judges} (additional uniform-judge passes), which are higher cost given per-pass compute.

\textbf{Counterfactual validity (deep dive).} CSS as computed treats every text-level mutation as a clinically meaningful counterfactual, but three failure modes can produce false low scores: (i) \emph{semantic no-op} mutations where regex-level text changes do not change clinical meaning; (ii) \emph{incomplete propagation} where one part of the chart is mutated while other parts continue to imply the original fact (e.g., deleting a Procedure block while encounter notes still refer to ``post-op''); (iii) \emph{medical incoherence} where the inserted scenario is medically impossible (e.g., curative resection on a metastatic patient), surfaced through human validation as $29/49$ Family D rows. A model that correctly refuses to update on any of these is scored 0.0 under the pre-registered rule. The camera-ready filter tightening and refusal-credit branch above address (iii) directly; (i) and (ii) require the semantic-validity audits in future work.

\textbf{Pre-registration tradeoff.} Pre-registration commits us to a scoring rule before observing outputs, which protects against post-hoc cherry picking but cannot distinguish causal-sensitivity failure from correct refusal of incoherent or non-actionable inputs. We report results under the original rule and flag the medical-incoherence fraction transparently in \S\ref{sec:human_validation}; quantifying semantic no-ops and incomplete-propagation fractions requires the semantic-validity audits listed above.

\end{document}